\pdfoutput=1

\PassOptionsToPackage{hyphens}{url}
\documentclass[11pt]{article}

\usepackage{acl}

\hypersetup{
  pdftitle={A Mechanistic Study of AI-Text Detection Neurons in Frozen BERT: Sparse Probing and Activation Patching on RAID},
  pdfauthor={Paweł Blicharz, Miłosz Grunwald}
}

\usepackage{times}
\usepackage{latexsym}
\usepackage[T1]{fontenc}
\usepackage[utf8]{inputenc}
\usepackage{microtype}

\usepackage{booktabs}     
\usepackage{multirow}
\usepackage{amsmath}
\usepackage{amssymb}
\usepackage{graphicx}
\usepackage{xcolor}
\usepackage{xspace}
\usepackage{tikz}
\usetikzlibrary{positioning, arrows.meta, calc}
\usepackage{placeins}  

\newcommand{\bertbase}{BERT-base-uncased\xspace}
\newcommand{\nselected}{n_{\text{sel}}\xspace}

\title{A Mechanistic Study of AI-Text Detection Neurons in Frozen BERT:\\
Sparse Probing and Activation Patching on RAID}

\author{
  Paweł Blicharz \\
  Gradient PG \\
  Gdańsk University of Technology \\
  \And
  Miłosz Grunwald \\
  Gradient PG \\
  Gdańsk University of Technology \\
}

\begin{document}
\maketitle

\begin{abstract}
AI-generated text detectors achieve high accuracy on standard benchmarks, yet the internal representations that drive these predictions remain poorly understood. We study which neurons in a \textit{frozen} \bertbase encoder support AI-text detection, using the RAID benchmark across six generators spanning pure-base and instruction-tuned models. We apply the L1$\to$L2 sparse-probing protocol of \citet{gurnee2023haystack} to all 9{,}216 CLS hidden-state dimensions (12 layers $\times$ 768), which we call \textit{neurons}. The procedure recovers a stable set of under 1\% of neurons per generator, consistent across folds and seeds; a probe restricted to that set retains most of the full-feature detection accuracy. Bidirectional activation patching confirms this set's causal relevance: in both directions it flips predictions an order of magnitude more often than size-matched random sets. Mean-ablating the same neurons leaves accuracy largely intact; the signal is therefore redundantly distributed. Cross-generator analysis reveals a bipartite structure: instruction-tuned generators concentrate 30--36\% of stable neurons in BERT's final layer while both base generators fall below 14\%, consistent with a layer-12 footprint of post-training alignment. Leave-one-family-out evaluation shows the selected neurons retain 86--94\% of the full-feature ceiling on unseen generator families, so a detector can operate on a small fixed subspace without re-identifying neurons per generator.
\end{abstract}

\section{Introduction}
\label{sec:introduction}
\begin{figure*}[!t]
\centering
\resizebox{\textwidth}{!}{%
\begin{tikzpicture}[
  >=Stealth, thick,
  box/.style={
    draw, rounded corners=3pt, fill=blue!6,
    minimum width=2.1cm, minimum height=0.78cm,
    align=center, font=\small},
  frozen/.style={
    draw, dashed, rounded corners=3pt, fill=gray!11,
    minimum width=2.1cm, minimum height=0.78cm,
    align=center, font=\small},
  outcome/.style={
    draw, rounded corners=3pt, fill=teal!9,
    minimum width=2.4cm, minimum height=0.78cm,
    align=center, font=\small},
  note/.style={font=\footnotesize, text=black!65, align=center},
]
\node[box]                        (inp)  {Input texts\\{\footnotesize human / AI}};
\node[frozen, right=1.05cm of inp]  (bert) {\bertbase\\{\footnotesize (frozen)}};
\node[box,    right=1.05cm of bert] (z)    {$\mathbf{z}\!\in\!\mathbb{R}^{9{,}216}$\\{\footnotesize CLS, layers 1--12}};
\node[box,    right=1.05cm of z]    (l1)   {L1 probe\\{\footnotesize $C\!=\!0.005$}};
\node[box,    right=2.2cm of l1]    (S)    {Stable set $\mathcal{S}^{*}$\\{\footnotesize 45--62 neurons}};
\draw[->] (inp)  -- (bert);
\draw[->] (bert) -- (z);
\draw[->] (z)    -- (l1);
\draw[->] (l1)   -- node[above=4pt, note] {${\geq}80\%$ of 15 cells} (S);
\node[note, above=0.85cm of l1] {5-fold $\times$ 3 seeds = 15 cells};
\coordinate (fork) at ($(S.south)+(0,-0.5)$);
\draw[-] (S.south) -- (fork);
\node[box, below left=1.55cm and 2.1cm of S] (ablbox)
  {Mean ablation\\{\footnotesize $z_{\mathcal{S}^{*}} \leftarrow \bar{\mu}_{\mathrm{train}}$}};
\node[frozen,  below=0.55cm of ablbox] (l2a)
  {L2 eval probe\\{\footnotesize (frozen)}};
\node[outcome, below=0.55cm of l2a]   (rA)
  {Acc.\ drop $\leq\!0.11$\,pp\\{\footnotesize 5/6; 1.06\,pp Cohere}\\{\footnotesize \S\ref{sec:results-necessity}}};
\draw[->] (fork) -| (ablbox.north);
\draw[->] (ablbox) -- (l2a);
\draw[->] (l2a)    -- (rA);
\node[note, above=0.18cm of ablbox] {\textit{necessity test}};
\node[box, below right=1.55cm and 0.55cm of S] (patbox)
  {Activation patching\\{\footnotesize bidirectional}};
\node[frozen,  below=0.55cm of patbox] (l2p)
  {L2 eval probe\\{\footnotesize (frozen)}};
\node[outcome, below=0.55cm of l2p]   (rP)
  {Fwd: 1--8\% (10--16$\times$)\\{\footnotesize Rev: 0.65--5.74\% (7--12$\times$)}\\{\footnotesize \S\ref{sec:results-patching}}};
\draw[->] (fork) -| (patbox.north);
\draw[->] (patbox) -- (l2p);
\draw[->] (l2p)    -- (rP);
\node[note, above=0.18cm of patbox] {\textit{sufficiency test}};
\end{tikzpicture}%
}
\caption{Overview of the experimental pipeline. Frozen \bertbase CLS activations are concatenated across all 12 layers, yielding $\mathbf{z}\in\mathbb{R}^{9{,}216}$. An L1-regularised probe selects a candidate set per (fold, seed) cell; neurons stable in ${\geq}80\%$ of all 15 cells become the \textit{stable set}~$\mathcal{S}^{*}$. An independent L2 probe trained on all 9{,}216 neurons is held frozen as the decision function for both experiments. \textbf{Left branch} (\S\ref{sec:results-necessity}): mean ablation. \textbf{Right branch} (\S\ref{sec:results-patching}): bidirectional activation patching, forward (AI~$\to$~human) and reverse (human~$\to$~AI). Numerical results are reported in the cited sections.}
\label{fig:pipeline}
\end{figure*}
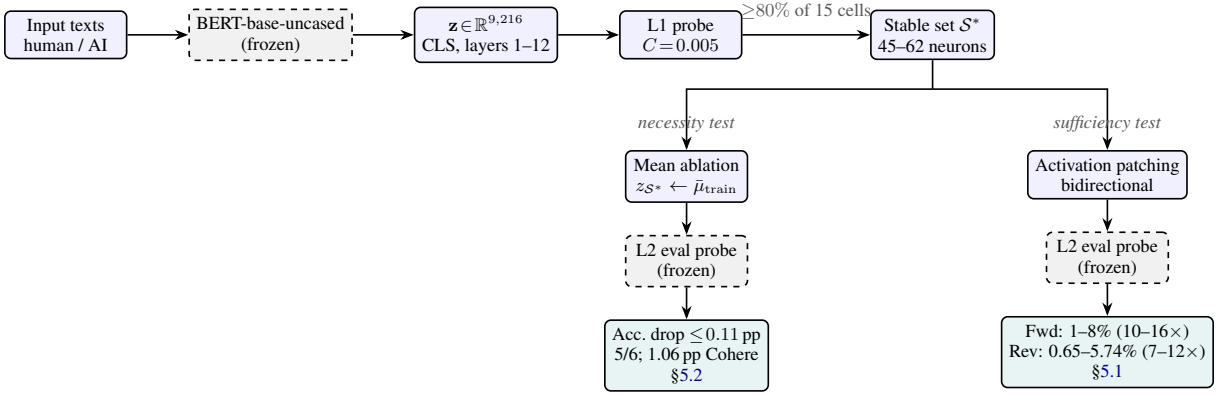

The rapid development of Large Language Models (LLMs) has made AI-generated text increasingly difficult to distinguish from human writing~\cite{dugan2024raid}. This poses a significant challenge across domains ranging from academic integrity to social media moderation and journalistic authenticity. Although detection accuracy has improved substantially, with encoder-based classifiers reaching high performance on established benchmarks~\cite{dugan2024raid}, a fundamental question remains unanswered: \textit{What internal representations drive these predictions?} Mechanistic interpretability~\citep{elhage2021framework,bereska2024mechanistic} has seen little application to AI-text detection, leaving the field without a causal explanation of how detection works.

Existing interpretability work on AI-text detection falls into five strands: (a)~\textit{black-box feature attribution} via SHAP~\citep{lundberg2017shap} or LIME~\citep{ribeiro2016lime}, which identifies predictive tokens without inspecting model internals~\cite{pudasaini2026why}; (b)~\textit{confounder removal}, which suppresses neurons correlated with domain or style artefacts in fine-tuned encoders to improve out-of-distribution accuracy~\cite{borile2025confounding}; (c)~\textit{sparse autoencoder feature analysis} applied to decoder-based models~\cite{kuznetsov2025sae}; (d)~\textit{probing on fine-tuned detectors}, which reveals reliance on narrow lexical or structural cues~\citep{choi2025probing}; and (e)~\textit{topological analysis} of attention maps, characterising surface and syntactic properties that differentiate AI from human text~\citep{kushnareva2021topology}. These approaches share a fundamental limitation documented across neuron-level analysis broadly~\citep{sajjad2022survey}: they are \textit{observational}. They identify neurons that correlate with detector behaviour or remove neurons that hurt generalisation, but none directly tests whether a small set of internal representations is \textit{causally sufficient} to produce AI-text predictions.

We address this gap with a causal intervention study on a \textit{frozen} \bertbase; Figure~\ref{fig:pipeline} summarises the experimental pipeline. Rather than fine-tuning the model for detection, we train sparse linear probes on pretrained CLS-token activations concatenated across all 12 layers (9{,}216 neurons total), following the sparse-probing protocol of \citet{gurnee2023haystack}. An L1-regularised logistic regression selects a per-cell candidate set; neurons that survive across folds and seeds form the aggregated stable set $\mathcal{S}^{*}$. An independent L2 probe serves as the frozen evaluator for all subsequent experiments. We then apply two causal interventions to the same selected neuron set: \textit{mean ablation}, which replaces these neurons with their training-set means to test necessity; and \textit{same-domain bidirectional activation patching}, run in both the forward direction (AI donor $\to$ human target, testing sufficiency) and the reverse direction (human donor $\to$ AI target, testing partial necessity on-manifold). The intervention orientation is the key distinction from prior work: rather than suppressing suspected confounders, we inject donor activations and measure whether predictions flip.

Across six generators spanning two pure-base and four instruction-tuned families, bidirectional activation patching shows the selected neurons are sufficient to drive the probe's AI predictions (forward, 10--16$\times$ above random) and partially necessary to preserve them (reverse, 7--12$\times$ above random). Mean-ablation of the same neurons leaves accuracy largely intact. The detection signal is therefore \textit{redundantly distributed}: easy to inject, hard to erase. Across generators, the stable sets are generator-specific yet overlap $\sim$24$\times$ above chance, with a layer-12 concentration in instruction-tuned generators not present in either base model. Leave-one-family-out evaluation confirms that the selected neurons retain most of the full-feature detection signal on entirely unseen generator families.

Our work is most directly related to \citet{borile2025confounding}, who suppress confounding neurons in a fine-tuned encoder to improve out-of-distribution detection, an observational approach that identifies which neurons hurt generalisation rather than which neurons causally drive predictions. More broadly, \citet{antverg2022pitfalls} show that individual-neuron analysis in language models is prone to false positives without causal validation (a pitfall observed in recent neuron-level analysis of AI-generated text~\citep{enkhbayar2025atomic}). We address the lack of a sufficiency test for AI-detection neurons directly via activation patching, and discuss the relationship to this and other prior work in Section~\ref{sec:related}.

\paragraph{Contributions.}
We make four contributions:
\begin{enumerate}
\item A sparse-probing characterisation of AI-detection signal in frozen \bertbase across six RAID generators. Stable sets of $<$1\% of CLS dimensions are sufficient for near-ceiling detection accuracy with high cross-fold stability.
\item The first application of bidirectional activation patching to AI-text detection neurons. Forward patching establishes that the selected neurons are sufficient to drive the probe's AI predictions (10--16$\times$ above random); reverse patching establishes that they are partially necessary to preserve them (7--12$\times$ above random).
\item A cross-generator analysis. Generator-specific stable sets nonetheless overlap ${\sim}24\times$ above chance, with a layer-12 footprint concentrated in instruction-tuned generators and absent in both pure-base models.
\item A leave-one-family-out (LOGO) evaluation. The selected neurons retain 86--94\% of the full-feature detection signal on entirely unseen generator families.
\end{enumerate}

\section{Related Work}
\label{sec:related}

\subsection{AI-Generated Text Detection}

Supervised fine-tuned classifiers~\citep{solaiman2019release, zellers2019grover} achieve high in-distribution accuracy but degrade substantially on unseen generators or domains. Zero-shot methods such as DetectGPT~\citep{mitchell2023detectgpt} and Binoculars~\citep{hans2024binoculars} exploit log-probability statistics without labelled data but require whitebox or API access to a reference model. Systematic benchmarking via RAID~\citep{dugan2024raid}, spanning 11 generators and 8 domains, confirms that no single method generalises reliably across generators or domains. This motivates our focus on what internal representations \textit{drive} detection, rather than improving detection accuracy itself.

\subsection{Interpretability of Text Detectors}

The simplest form of interpretability for AI-text detectors uses input-level attribution. \citet{pudasaini2026why} apply SHAP to explain classifier decisions, finding that the most predictive tokens diverge substantially across domains. This approach characterises the model's decision surface but cannot speak to what internal representations underlie it.

More recent work examines representations inside the encoder. \citet{borile2025confounding} identify neurons in a fine-tuned BERT encoder whose activations correlate with domain or style artefacts rather than AI-authorship signal, and suppress them to improve out-of-distribution accuracy. \citet{kuznetsov2024restricted} take a geometric approach, removing a learned embedding subspace to produce a detector more robust to lexical variation. Both works operate on fine-tuned models and ask which neurons hurt OOD generalisation, not which neurons are sufficient to drive predictions. Our work differs from \citet{borile2025confounding} on four axes: (i)~we use a \textit{frozen} rather than fine-tuned encoder; (ii)~we probe CLS-token representations rather than layer-specific FFN sublayer activations; (iii)~we perform both ablation to measure \textit{necessity} and activation patching to test causal \textit{sufficiency}, rather than ablating to identify and suppress confounders; and (iv)~we evaluate on RAID rather than DAIGT, M4~\citep{wang2024m4}, or HC3~\citep{guo2023hc3}. The terminological consequence is that \citet{borile2025confounding} identify confounding neurons in the MLP intermediate layers of a fine-tuned BERT detector ($H = 3{,}072$ units per layer); we instead probe the residual-stream output at the CLS position (768 dimensions per layer), following the broader \citet{sajjad2022survey} convention.

\citet{kuznetsov2025sae} apply sparse autoencoders to Gemma-2 decoder activations, extracting semantically interpretable features correlated with AI authorship. This work is complementary but distinct. The model class (autoregressive decoder), representation type (token-position activations rather than CLS pooling), and analysis method (SAE reconstruction) all differ from ours. The approach also remains observational: features are identified by correlation, not by causal intervention. None of the above provides direct evidence that a small identified set of neurons is \textit{sufficient} to produce AI-text predictions when injected into human-text activations; that is the question we address.

\subsection{Sparse Probing and Activation Patching}

Earlier work identified task-relevant neurons in pretrained models via correlation-based selection, and mapped salient neurons to linguistic properties across layers and model families~\citep{durrani2023salient}. Sparse probing uses L1-regularised linear classifiers to identify a small set of neurons in a pretrained model that carry task-relevant information. \citet{gurnee2023haystack} introduced and systematically evaluated the L1-then-L2 protocol: an L1 probe selects candidate neurons, and an L2 probe re-fitted on those neurons provides unbiased accuracy estimates. Their study shows that individual neurons in GPT-2 and LLaMA encode grammatical, factual, and linguistic properties, and that the selected sets are stable across seeds and data subsets. We adopt this protocol directly, applying it to \bertbase CLS-token activations concatenated across all 12 layers.

Activation patching replaces internal representations in a target forward pass with those from a donor pass, then measures the effect on model outputs~\citep{vig2020causal,meng2022rome}. \citet{heimersheim2024patching} recommend donor patching with real inputs (rather than zero or mean vectors) to avoid out-of-distribution artefacts, and mean ablation as a conservative necessity baseline; we follow both. The closest methodological precedent for our flip-rate framing is \citet{ravindran2025adversarial} (donor patching, flip-rate metric; autoregressive decoder, safety alignment). Activation patching has also been applied to BERT cross-encoders for IR~\citep{lu2025pathway}, but to our knowledge no prior work applies it to encoder-based AI-text detection.

\section{Experimental Setup}
\label{sec:setup}

\subsection{Data: RAID}
\label{sec:setup-data}

RAID~\citep{dugan2024raid} provides human-written and machine-generated texts across 11 generators and 8 domains. We use six of the eight RAID domains: abstracts, books, news, Reddit posts, reviews, and Wikipedia. We exclude poetry and recipes, as both have very low sample counts in RAID and risk introducing genre-specific artefacts that confound cross-domain analysis. We select six generators spanning five model families: GPT-4~\citep{openai2023gpt4}, GPT-2~\citep{radford2019gpt2}, MPT-30B~\citep{mosaicml2023mpt}, Mistral-7B-Instruct-v0.1~\citep{jiang2023mistral}, LLaMA-2-Chat~\citep{touvron2023llama2}, and Cohere-Command.\footnote{Cohere Command does not have a canonical reference paper; see the model documentation at \url{https://cohere.com/command}.} This covers two pure-base generators (GPT-2, MPT-30B), one SFT-only generator (Mistral-7B-Instruct-v0.1, fine-tuned on instruction data without RLHF~\citep{jiang2023mistral}), and three SFT+RLHF generators (GPT-4, LLaMA-2-Chat, Cohere-Command). The six were chosen to span two axes rather than to cover RAID exhaustively: model family (one generator per family) and post-training regime (every regime present in the benchmark). This design enables the base vs.\ instruction-tuned comparison in §\ref{sec:results-crossgen}. The remaining five RAID generators (GPT-3, ChatGPT, Mistral-7B base, MPT-Chat, Cohere-Chat) are same-family duplicates; none contributes an architecture family or a post-training regime that the six do not already cover. All 11 generators are used in the leave-one-family-out evaluation of §\ref{sec:logo}. We subsample 7{,}500 examples per generator (3{,}750 human, 3{,}750 AI) balanced across domains, yielding 45{,}000 samples in total.

\subsection{Model and Activations}
\label{sec:setup-model}

We use \bertbase~\citep{devlin2019bert} with all weights frozen throughout. For each input text we extract the CLS-token hidden state at every transformer layer $l \in \{1, \ldots, 12\}$, giving a per-layer vector $\mathbf{h}^{(l)} \in \mathbb{R}^{768}$. The full hidden-state representation is the concatenation
\[
  \mathbf{z} = \bigl[\mathbf{h}^{(1)};\, \ldots;\, \mathbf{h}^{(12)}\bigr] \in \mathbb{R}^{9{,}216}.
\]
We refer to individual dimensions of $\mathbf{z}$ as \textit{neurons}; neuron index $i$ resides in layer $\lfloor i / 768 \rfloor + 1$ and position $i \bmod 768$ within that layer.\footnote{The term \textit{neuron} carries two meanings in the literature. The NLP probing tradition~\citep{dalvi2019grain,sajjad2022survey,durrani2023salient} uses it for any scalar coordinate of a hidden representation; some mechanistic-interpretability work~\citep{elhage2021framework,gurnee2023haystack} restricts it to MLP post-activation units, where the elementwise nonlinearity yields a privileged basis absent from the residual stream. We follow the former; individual residual-stream coordinates in BERT are known to carry meaningful signal on their own~\citep{kovaleva2021bertbusters,timkey2021rogue}.} Using the frozen pretrained model (rather than a fine-tuned detector) ensures that any informative neurons we identify are present in general-purpose BERT representations, not artefacts of task-specific fine-tuning.

\subsection{Protocol: L1 Selection, L2 Evaluation}
\label{sec:setup-protocol}

Following \citet{gurnee2023haystack}, for each (seed, fold) cell we fit two independent probes on the training split. (1)~\textit{L1 selection probe}: L1-regularised logistic regression on all 9{,}216 neurons, fitted with scikit-learn's~\citep{scikit-learn} \texttt{liblinear} solver at inverse regularisation strength $C = 0.005$ (smaller $C$ $\Rightarrow$ stronger L1 penalty; see Appendix~\ref{app:pipeline} for the exact loss). Neurons with non-zero coefficients form the per-cell \textit{candidate set} $\mathcal{S}$; neurons appearing in $\mathcal{S}$ in at least 80\% of the 15 cells form the aggregated \textit{stable set} $\mathcal{S}^{*}$. Interventions (patching, ablation) target the per-fold $\mathcal{S}$; $\mathcal{S}^{*}$ is reserved for cross-generator analyses (§\ref{sec:results-crossgen}) and LOGO. (2)~\textit{L2 evaluation probe}: L2-regularised logistic regression (lbfgs) on the full $\mathbf{z}$. This probe is held frozen for all downstream experiments to ensure that intervention effects are measured against a fixed decision function.

We use 5-fold cross-validation stratified by (label $\times$ domain) with seeds $\{42, 123, 456\}$, yielding 15 cells per (generator, experiment). All reported metrics are 15-cell means $\pm$ standard deviation unless otherwise noted; per-cell values are tabulated in Appendix~\ref{app:per-cell}.

\paragraph{Hyperparameter selection.}
\label{sec:setup-stability}%
$C = 0.005$ and $N = 7{,}500$ were chosen via a multi-generator stability sweep run prior to any results analysis, maximising the minimum-over-generators pairwise Jaccard subject to $\geq 50$ neurons per draw~\citep{gurnee2023haystack}. The chosen operating point achieves a minimum Jaccard of 0.676 (GPT-2 bottleneck; other five generators 0.707--0.751); full grid available in Appendix~\ref{app:stability}.

\section{Sparse Detection Neurons in Frozen BERT}
\label{sec:results-sparse}

Table~\ref{tab:sparse-summary} reports sparse-probe results across all six generators. Between 45 and 62 neurons (0.49--0.67\% of the 9{,}216-dimension CLS representation) are stable across at least 12 of the 15 cells, yet the L2 evaluation probe achieves 90.5--98.8\% validation accuracy (AUC-ROC 0.968--0.999). A probe restricted to the stable set alone achieves 86.5--97.2\% (see Appendix~\ref{app:restricted}); the stable neurons are therefore themselves sufficient for most of the detection accuracy. Selection is highly stable: pairwise fold Jaccard ranges 0.67--0.76 across 105 fold-pairs per generator, compared to a random-null expectation of $\sim$0.003--0.005 for sets of this size drawn from 9{,}216 neurons, placing within-generator stability at 150--230$\times$ above chance. The selection pool is narrow: between 91 neurons ever selected (LLaMA) and 146 (GPT-2). The probe consistently selects from a small region of the feature space. Layer-by-layer neuron distributions and the base/instruction-tuned asymmetry are examined in Section~\ref{sec:results-crossgen}.

\begin{table}[t]
\centering
\resizebox{\columnwidth}{!}{%
\begin{tabular}{lrrrrr}
\toprule
Generator & Val acc & AUC & $\nselected$ & Jaccard & $n_{\text{stable}}$ / $n_{\text{obs}}$ \\
\midrule
GPT-4   & $0.988 \pm 0.003$ & $0.999 \pm 0.000$ & $74.1 \pm 2.9$ & $0.763 \pm 0.037$ & 60 / 106 \\
GPT-2   & $0.959 \pm 0.003$ & $0.992 \pm 0.001$ & $84.7 \pm 4.1$ & $0.670 \pm 0.042$ & 60 / 146 \\
MPT     & $0.972 \pm 0.003$ & $0.994 \pm 0.001$ & $61.3 \pm 3.1$ & $0.740 \pm 0.036$ & 52 / 94  \\
Mistral & $0.971 \pm 0.004$ & $0.995 \pm 0.001$ & $57.8 \pm 2.9$ & $0.724 \pm 0.046$ & 45 / 99  \\
LLaMA   & $0.986 \pm 0.003$ & $0.998 \pm 0.001$ & $63.6 \pm 1.7$ & $0.756 \pm 0.040$ & 48 / 91  \\
Cohere  & $0.905 \pm 0.008$ & $0.968 \pm 0.003$ & $72.3 \pm 3.1$ & $0.749 \pm 0.039$ & 62 / 113 \\
\bottomrule
\end{tabular}%
}
\caption{Sparse-probe results across six generators: 15-cell means $\pm$ std (5 folds $\times$ 3 seeds). Val acc: L2 evaluation probe accuracy (all 9{,}216 neurons). AUC: area under the ROC curve. $\nselected$: mean candidate set size per cell. Jaccard: mean pairwise fold Jaccard over all 105 fold-pairs (C(15,2)), computed from per-fold selection sets. $n_{\text{stable}}$: neurons in $\geq$80\% of cells; $n_{\text{obs}}$: total distinct neurons selected at least once. The six generators span one model family each and all post-training regimes present in RAID (two pure-base, one SFT-only, three SFT+RLHF); the remaining five RAID generators are same-family, same-regime duplicates (§\ref{sec:setup-data}).}
\label{tab:sparse-summary}
\end{table}

\section{Causal Role of the Selected Neurons}
\label{sec:results-causal}

Having identified a sparse set of neurons, we ask two complementary causal questions: are these neurons \textit{sufficient} to drive the probe's AI predictions, and are they \textit{necessary} to preserve them? Section~\ref{sec:results-patching} addresses both via bidirectional activation patching: transplanting AI activations into human targets (forward; sufficiency) and human activations into AI targets (reverse; partial necessity). Section~\ref{sec:results-necessity} complements this with mean ablation. We provide a restricted-probe localisation analysis in Appendix~\ref{app:restricted}; it reinforces the base/instruction-tuned split.

\subsection{Causal Role via Bidirectional Activation Patching}
\label{sec:results-patching}

\subsubsection{Patching Protocol}
\label{sec:patching-protocol}

We run patching \textbf{bidirectionally} to assess both causal directions: (i)~\textbf{forward} (AI donor $\to$ human target): transplanting the selected neurons' activations from AI samples into human targets, testing whether the selected set is \textit{sufficient} to drive the probe's AI predictions; and (ii)~\textbf{reverse} (human donor $\to$ AI target): transplanting human activations into AI targets, testing whether the selected set is partially \textit{necessary} for maintaining AI predictions.

For each (seed, fold) cell and direction we identify \textit{valid target} samples: test-fold samples of the target class that share a domain with at least one test-fold donor sample. For each of $n_{\text{shuffles}} = 20$ random within-domain donor permutations we construct a patched activation by replacing the selected-neuron activations of the target with those of the donor ($\tilde{\mathbf{z}}_{\mathcal{S}} \!\leftarrow\! \mathbf{z}^{\text{donor}}_{\mathcal{S}}$), leaving non-selected neurons unchanged. A \textit{flip} occurs when the frozen evaluation probe's predicted class changes after patching (human $\to$ AI in the forward direction; AI $\to$ human in the reverse). The flip rate is the fraction of valid target samples that flip, averaged over the 20 donor permutations and then over the 15 cells. Donor sampling is restricted to the same domain as the target, following the on-manifold recommendation of \citet{heimersheim2024patching} to avoid injecting out-of-distribution activations.

The \textit{random-$k$ baseline} draws 20 random neuron subsets of size $k$ and runs the identical procedure. This serves as a null that controls for the number of patched dimensions. We sweep $k \in \{1, 5, 10, 20, 50,\ \text{full}\}$ for both the selected-neuron patching and the random-k baseline. $k = \text{full}$ refers to the per-fold L1-selected set $\mathcal{S}$ (53--92 neurons depending on generator and fold; see §\ref{sec:setup-protocol} for the per-fold $\mathcal{S}$ vs aggregated $\mathcal{S}^{*}$ distinction).

\subsubsection{Flip-Rate Results (Forward and Reverse)}
\label{sec:patching-results}

\begin{figure}[t]
\centering
\includegraphics[width=\columnwidth]{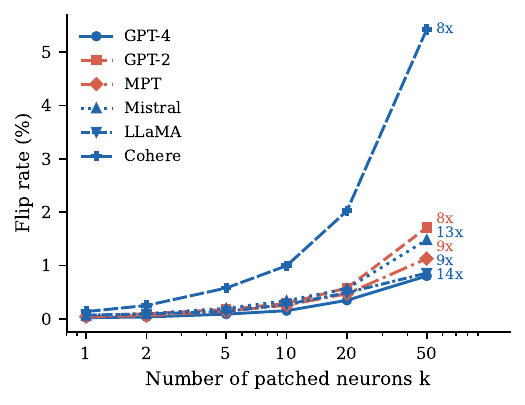}
\caption{Forward flip rate (\%) vs.\ number of patched neurons $k$ on a log-scale x-axis, averaged over 15 cells (5 folds $\times$ 3 seeds). Instruction-tuned generators are shown in blue, base generators in red. Numbers to the right of each line are selected/random ratios at $k=50$ (8--14$\times$); headline ratios at $k = \text{full}$ (the per-fold L1-selected set, 53--92 neurons) are 10--16$\times$ across generators (Table~\ref{tab:patching-sweep}).}
\label{fig:flip-rate}
\end{figure}

\paragraph{Forward (AI $\to$ human).}
Flip rates at $k = \text{full}$ are 7--18$\times$ higher than at $k = 5$ across all generators (Figure~\ref{fig:flip-rate}; full sweep in Table~\ref{tab:patching-sweep}, Appendix~\ref{app:patching-sweep}; per-domain breakdown in Appendix~\ref{app:patching-perdomain}), indicating that the detection signal is distributed across the selected set rather than concentrated in a small subset.

\begin{table}[t]
\centering
\small
\begin{tabular}{lrrr}
\toprule
Generator & Selected (\%) & Random-$k$ (\%) & Ratio \\
\midrule
GPT-4   & $1.23_{\pm 0.33}$ & $0.08_{\pm 0.07}$ & $16\times$ \\
GPT-2   & $3.13_{\pm 0.54}$ & $0.32_{\pm 0.09}$ & $10\times$ \\
MPT     & $1.52_{\pm 0.21}$ & $0.14_{\pm 0.08}$ & $11\times$ \\
Mistral & $1.73_{\pm 0.27}$ & $0.12_{\pm 0.08}$ & $14\times$ \\
LLaMA   & $1.07_{\pm 0.37}$ & $0.11_{\pm 0.07}$ & $10\times$ \\
Cohere  & $8.15_{\pm 1.27}$ & $0.79_{\pm 0.16}$ & $10\times$ \\
\bottomrule
\end{tabular}
\caption{Forward-direction (AI $\to$ human) selected and random-$k$ flip rates (\%) at $k = \text{full}$ (15-cell means $\pm$ std). Ratio = selected / random (rounded to nearest integer; precise values 9.7--15.7$\times$). Full $k$-sweep in Appendix~\ref{app:patching-sweep}. Reverse-direction (human $\to$ AI) results are reported in the text.}
\label{tab:patching-full}
\end{table}

Generators nearer the decision boundary flip more: Cohere flips most readily (8.15\%, lowest baseline at 90.5\%), LLaMA least (1.07\%, baseline 98.6\%) (Table~\ref{tab:patching-full}).

\paragraph{Reverse (human $\to$ AI).}
Transplanting human-donor activations into AI-target samples tests whether the selected neurons are partially necessary for maintaining AI-class predictions. At $k = \text{full}$, 0.65--5.74\% of AI samples were reclassified as human (per-generator breakdown in Table~\ref{tab:patching-sweep}, Appendix~\ref{app:patching-sweep}), consistently 7--12$\times$ above the random-neuron baseline. Reverse flip rates are lower than forward (0.65--5.74\% vs.\ 1.07--8.15\%): the remaining 9{,}000$+$ unpatched neurons carry enough redundant AI signal to partially preserve the prediction even when the selected set is overwritten. The same-domain donor constraint eliminates domain shift by construction; mean-ablation caveats are discussed in §\ref{sec:results-necessity}. Appendix~\ref{app:worked-example} reports one patched sample in full. Absolute flip rates are low because a flip requires overturning a probe that reads all 9{,}216 neurons while fewer than 1\% of them are edited, so the informative quantity is the ratio to the size-matched random-$k$ baseline (10--16$\times$ forward, 7--12$\times$ reverse) rather than the raw percentage. That the selected neurons carry the detection signal, rather than merely perturbing it, is established separately by the restricted probe, which reaches 86.5--97.2\% accuracy on those neurons alone (Appendix~\ref{app:restricted}).

\paragraph{Domain structure.}
Flip rates are not uniform across RAID domains, and the pattern differs by
generator: Cohere is uniformly high across all six domains while GPT-2 is
elevated on news, Reddit, and Wikipedia relative to books and reviews.
Appendix~\ref{app:patching-perdomain} gives the per-domain breakdown for every
generator (Table~\ref{tab:perdomain}) together with a dispersion analysis
showing that domain sensitivity does not follow the base/instruction-tuned
split that organises the layer distributions (§\ref{sec:crossgen-layers}).

\subsection{Necessity via Mean Ablation}
\label{sec:results-necessity}

To complement the sufficiency test we ask the converse question: is the selected set \textit{necessary}? We answer this via mean ablation, replacing the selected neurons in the test-fold activations with their training-fold mean (computed on the training split to avoid leakage) and re-evaluating the frozen L2 probe. Table~\ref{tab:ablation-full} reports accuracy drop at $k = \text{full}$ alongside the random-$k$ baseline.

For five of six generators the result is negative. At $k = \text{full}$, ablating the selected set causes accuracy to drop by only 0.03--0.11\,pp, within noise of the 0.00--0.06\,pp random-$k$ baseline; the drops for GPT-4, GPT-2, MPT, and Mistral are within one standard deviation of zero (Table~\ref{tab:ablation-full}), so no reliable necessity is detectable for these generators. The selected neurons are not individually necessary: the L2 evaluation probe, which uses all 9{,}216 neurons, retains thousands of redundant correlates that compensate for the removed subset.

Cohere is the exception. Its full-set ablation drop is 1.06\,$\pm$\,0.52\,pp (well outside the 0.06\,pp random baseline and the per-cell noise floor), and the drop grows monotonically from $k = 5$ upward, unlike the other five generators where the curve is flat. Its other signals point the same way: lowest baseline accuracy, smallest mean-difference norm, and largest probe weight norm (CAV diagnostics, Appendix~\ref{app:cav}; AUC-ROC-vs-L1 reversal in Appendix~\ref{app:auc-l1}). Together these point to a detection representation that is \textit{partially} localised rather than fully distributed.

We treat the necessity result as a conservative lower bound: mean ablation moves activations off-manifold~\citep{heimersheim2024patching}, and the on-manifold reverse-patching test (§\ref{sec:patching-results}) confirms partial necessity. If L1 had selected only high-variance correlates, donor injection would not flip human predictions at $10\text{--}16\times$ random. The selected set is one sufficient subspace among others, singled out by stable recovery across folds, seeds, and (via LOGO) generator families.

\begin{table}[t]
\centering
\resizebox{\columnwidth}{!}{%
\begin{tabular}{lrrr}
\toprule
Generator & Base acc & Sel.\ drop & Rand.\ drop \\
\midrule
GPT-4   & 0.988 & $+0.03 \pm 0.12$\,pp & $+0.01$\,pp \\
GPT-2   & 0.959 & $+0.11 \pm 0.31$\,pp & $+0.00$\,pp \\
MPT     & 0.972 & $+0.05 \pm 0.11$\,pp & $+0.00$\,pp \\
Mistral & 0.971 & $+0.10 \pm 0.21$\,pp & $+0.00$\,pp \\
LLaMA   & 0.986 & $+0.11 \pm 0.14$\,pp & $+0.02$\,pp \\
Cohere  & 0.905 & $\mathbf{+1.06 \pm 0.52}$\,pp & $+0.06$\,pp \\
\bottomrule
\end{tabular}%
}
\caption{Mean-ablation accuracy drop at $k = \text{full}$, 15-cell means $\pm$ std (positive = drop). Sel.\ drop: ablating the per-fold L1-selected set ($k = \text{full}$, 53--92 neurons); Rand.\ drop: mean over 20 random $k$-subsets. Cohere is the only generator substantially above the random baseline.}
\label{tab:ablation-full}
\end{table}

\section{Cross-Generator Representation Geometry}
\label{sec:results-crossgen}

\subsection{Layer Distribution of Stable Neurons}
\label{sec:crossgen-layers}

\begin{figure}[t]
\centering
\includegraphics[width=\columnwidth]{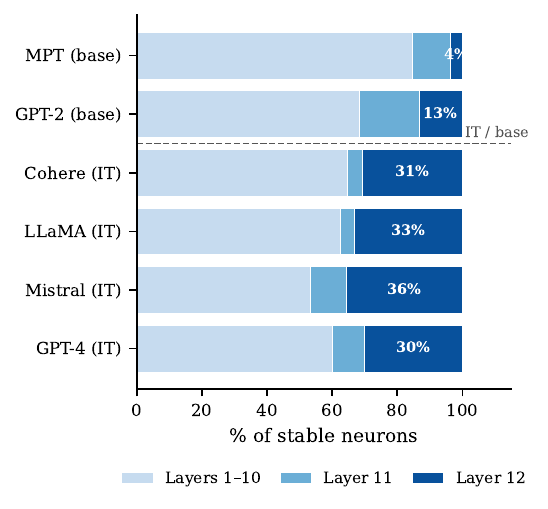}
\caption{Proportion of stable neurons in each layer group per generator. Layer~12 percentage annotated in white. Instruction-tuned generators (IT) concentrate 30--36\% of stable neurons in layer~12; base generators retain $\leq$14\%. Dashed line separates the two groups. The generators shown are one per RAID model family, selected to place both pure-base and both instruction-tuned post-training regimes on this axis; see §\ref{sec:setup-data} for the selection and exclusion criteria.}
\label{fig:layer-dist}
\end{figure}

Figure~\ref{fig:layer-dist} shows the layer-by-layer distribution of stable neurons across all six generators (per-layer counts in Appendix~\ref{app:layer-dist}). The four instruction-tuned generators and the two pure-base generators differ sharply. For GPT-4, Mistral, LLaMA, and Cohere, the single most populated layer is layer~12, which accounts for 30--36\% of each generator's stable set (16--19 neurons). Both pure-base generators are exceptions: GPT-2 peaks at layer~11 with 11 stable neurons (18.3\% of its stable set) and has only 13.3\% in layer~12; MPT peaks earlier with just 2 stable neurons in layer~12 (3.8\%) and distributes its signal more broadly across earlier layers (layers 1, 8, and 9 dominate).

Across both pure-base generators, the shared pattern is markedly low layer-12 concentration (3.8--13.3\%) versus any instruction-tuned generator (30--36\%). The L12-avoidance pattern is not a rigid L11-peaking rule (GPT-2 peaks at L11 while MPT peaks at L1/L8). On this evidence, BERT's layer~12 acts as a task-linked readout that instruction-tuned outputs activate more strongly than base-model outputs~\citep{tenney2019bert, rogers2020primer, geva2021ffn}. Because this reading rests on $N=2$ base generators (GPT-2, MPT) and describes the selected-neuron geometry rather than a mechanistic claim about RLHF/SFT-induced circuitry, we treat it as a working hypothesis (see \hyperref[sec:limitations]{Limitations}). Cohere's partial localisation (§\ref{sec:results-necessity}) fits this reading: an IT generator encoding the L12 signal at lower per-neuron magnitude, with the probe compensating via larger weights.

\subsection{Stable-Set Overlap Across Generators}
\label{sec:crossgen-jaccard}

Pairwise Jaccard similarity between stable sets (Table~\ref{tab:jaccard} in Appendix~\ref{app:jaccard}) ranges from 0.00 to 0.24. Stable sets are therefore generator-specific. The appropriate baseline is not zero but the expected Jaccard for two random sets of the same sizes drawn from 9{,}216 neurons, which is approximately 0.003 per pair. Relative to this null, the mean observed Jaccard of 0.073 is $\sim$24$\times$ above chance, indicating a partial shared substrate despite the generator-specific structure.

The data reveal a clear bipartite structure (Figure~\ref{fig:jaccard} in Appendix~\ref{app:jaccard}). Among instruction-tuned generators, pairwise Jaccard ranges 0.08--0.24 (27--86$\times$ chance), with the strongest pairs being Cohere--Mistral (86$\times$) and LLaMA--Mistral (64$\times$). Among the two pure-base generators, MPT and GPT-2 share 12 neurons (J\,=\,0.120, 40$\times$ chance), more than several instruction--instruction pairs, so the base models form a coherent sub-cluster. In contrast, every base-vs-instruction-tuned pair has near-zero overlap (J\,$\leq$\,0.05), with three pairs at exactly zero (full breakdown in Table~\ref{tab:jaccard}). This base-vs-instruction-tuned partition explains the bulk of the cross-generator geometry.

\subsection{A Cross-Generator Core Concentrated in Layer~12}
\label{sec:crossgen-core}

Neurons appearing in the stable sets of at least 3 of the 6 generators (minimal majority threshold) form a \textit{core} set of 17 neurons. Of these, 10 (59\%) reside in layer~12, far exceeding any individual generator's L12 share (3.8--36\%); the remaining 7 are scattered across layers 1, 2, 5, 7, 8, and 11. This concentration suggests BERT's layer~12 is the primary cross-generator site for AI-detection signal: a majority of neurons shared across half or more of the studied generators converge there, even though no individual generator devotes a majority of its own stable set to L12. Appendix~\ref{app:features} examines what these neurons respond to by correlating the highest-weight stable neurons with four surface text statistics; the strongest associations are moderate ($|\rho| = 0.41$--$0.54$) and vary in sign across generators, so the selected subspace does not reduce to a single lexical cue.

\section{Leave-One-Family-Out Generalisation}
\label{sec:logo}

The experiments above characterise stable neurons for each generator
\emph{independently}. A leave-one-family-out (LOGO) evaluation tests
whether neurons identified from \textit{other} generators transfer to a
held-out generator family never seen during neuron selection.

\paragraph{Protocol.}
We partition all 11 RAID generators into five families: \texttt{gpt}, \texttt{cohere}, \texttt{llama}, \texttt{mistral}, and \texttt{mpt}. For each held-out family we run the sparse-probe pipeline on the remaining generators and evaluate two probes on the held-out set: (i)~\textbf{Full L2} on all 9{,}216 neurons, and (ii)~\textbf{Sel\,L2} on only the stable neurons identified from training families. To handle the larger combined training pool, LOGO uses a tighter regularisation ($C = 0.001$) and caps each generator's contribution at 5{,}000 samples (compared to $C = 0.005$, $N = 7{,}500$ in the per-generator headline runs). Family sizes are unequal (1--4 generators per family, following RAID's taxonomy), so training pool size and test-set size vary across folds. The Sel/Full ratio in Table~\ref{tab:logo} normalises for this by expressing each family's held-out accuracy as a fraction of its own full-feature ceiling.

\paragraph{Results.}
Table~\ref{tab:logo} reports the outcome.
Full\,L2 generalises strongly in all five conditions (88.6--99.7\% accuracy),
showing that frozen BERT representations are themselves informative across
the RAID generator space. The Sel\,L2 column shows the main result:
91--125 stable neurons (at most 1.4\% of 9{,}216 neurons) achieve
76.3--93.7\% accuracy on generators from entirely held-out families.
Expressed as a fraction of the full-feature ceiling for each family,
the selected set retains 86--94\% of the full-feature signal (Table~\ref{tab:logo},
Sel/Full column).

The worst-case family is \texttt{cohere} (Sel\,L2 = 76.3\%, 86\% of ceiling), matching its low pairwise Jaccard with other generators; the best is \texttt{llama} (93.7\%, 94\%), which benefits from high overlap within the instruction-tuned group (§\ref{sec:crossgen-jaccard}). \texttt{mpt} and \texttt{mistral} fall in the middle tier (85.6--85.7\%).

We interpret the 6--14\,pp residual gap between Sel\,L2 and Full\,L2 as evidence that a cross-family core of neurons carries most of the detection signal that transfers, while the remainder of the full-feature ceiling is supplied by \textit{family-specific} neurons that the LOGO procedure cannot observe by construction. This tracks the bipartite Jaccard structure (§\ref{sec:crossgen-jaccard}), where each generator's stable set is dominated by a generator-specific periphery layered on a small shared substrate.

\paragraph{Practical implication.}
Because the stable neurons are identified without observing the held-out
family, a detector built on frozen BERT can operate on at most 1.4\% of the
representation and still generalise to generator families that were
unavailable when the neuron set was chosen, without re-running selection for
each new generator. Sel\,L2 is the leave-one-family-out counterpart of the
restricted probe of Appendix~\ref{app:restricted}, which reaches 86.5--97.2\%
when trained and tested on the same generator; the 76.3--93.7\% here is the
price of never seeing the target family. We report this as a property of the
representation rather than as a detection system; the
\hyperref[sec:ethics]{Ethical considerations} section discusses the symmetric
risk that the same localisation informs evasion.

\begin{figure}[t]
\centering
\includegraphics[width=\columnwidth]{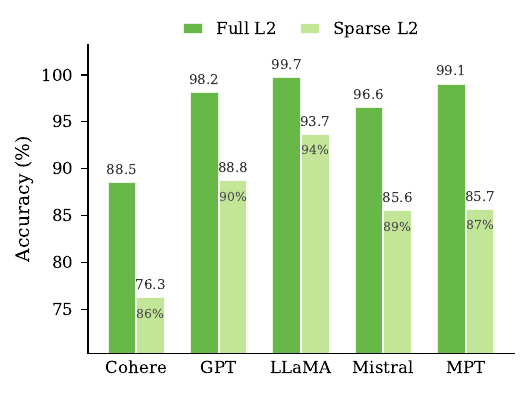}
\caption{Leave-one-family-out generalisation. For each held-out RAID family, the full-feature L2 probe (all 9{,}216 neurons) and the sparse L2 probe (only the 91--125 neurons selected from the four training families) are evaluated on the held-out family. Values above each bar are accuracies; the percentage inside each sparse bar is its retention relative to that family's own full-feature ceiling. The sparse probe never sees the held-out family during selection or fitting. Numerical values, including AUC, in Table~\ref{tab:logo}.}
\label{fig:logo}
\end{figure}

\begin{table}[t]
\centering
\resizebox{\columnwidth}{!}{%
\begin{tabular}{lrrrrrr}
\toprule
Family & $n_{\text{stable}}$ & Full\,L2 & Sel\,L2 & Sel/Full & Full AUC & Sel AUC \\
\midrule
cohere  & 119 & 88.6\% & 76.3\% & 86\% & 0.946 & 0.863 \\
gpt     &  91 & 98.2\% & 88.8\% & 90\% & 0.993 & 0.953 \\
llama   & 125 & 99.7\% & 93.7\% & 94\% & 0.999 & 0.986 \\
mistral & 106 & 96.6\% & 85.6\% & 89\% & 0.982 & 0.925 \\
mpt     & 110 & 99.1\% & 85.7\% & 87\% & 0.995 & 0.926 \\
\bottomrule
\end{tabular}%
}
\caption{Leave-one-family-out (LOGO) generalisation.
$n_{\text{stable}}$: stable neurons identified from the four training families.
Full\,L2: L2-probe accuracy using all 9{,}216 BERT neurons.
Sel\,L2: L2-probe accuracy restricted to $n_{\text{stable}}$ neurons.
Sel/Full: ratio expressing how much of the full-feature ceiling the selected
neurons retain.
Full AUC: area under the ROC curve (AUC) for the full-feature L2 probe.
Sel AUC: AUC-ROC for the Sel\,L2 probe, ranging from 0.863 (cohere) to 0.986 (llama).
All metrics are means over 3 seeds $\times$ 5 folds = 15 cells.
Test set sizes: 5{,}000 (llama, 1 generator) to 20{,}000 (gpt, 4 generators).}
\label{tab:logo}
\end{table}

\section{Conclusion}
\label{sec:conclusion}

An L1-regularised probe stably selects 45--62 neurons ($<$1\% of 9{,}216 CLS dimensions) across six RAID generators. Bidirectional activation patching shows these neurons are sufficient to drive the probe's AI predictions (forward, 10--16$\times$ above random) and partially necessary to preserve them (reverse, 7--12$\times$ above random); mean ablation costs $\leq$0.11\,pp on five of six generators (1.06\,pp on Cohere). The detection signal is \textit{easy to inject, hard to erase}: redundantly distributed yet concentrated enough to be sufficient~\citep{dalvi2020redundancy}.

The same data expose a bipartite structure: instruction-tuned generators concentrate stable neurons in BERT's final layer; base generators do not (see \hyperref[sec:limitations]{Limitations}). Leave-one-family-out evaluation confirms 86--94\% of the full-feature ceiling is retained on held-out families, with the residual gap reflecting family-specific neurons.

\section*{Limitations}
\label{sec:limitations}

\paragraph{Single encoder architecture.}
All experiments use frozen \bertbase. Whether the stable-neuron structure and patching effects extend to larger encoders (RoBERTa-large~\citep{liu2019roberta}, DeBERTa-v3~\citep{he2021debertav3}) or encoder-decoder architectures is an open question. The frozen setting is a deliberate design choice: it isolates what the pretrained representation already encodes, but it means results cannot be directly compared to fine-tuned detectors without controlling for the frozen-vs-fine-tuned distinction.

\paragraph{English-only evaluation.}
We evaluate exclusively on the English-language RAID benchmark. Whether the layer-12 footprint of post-training alignment, the bipartite base/instruction-tuned representation structure, and the redundancy patterns we observe generalise to languages with richer morphology, different syntactic structure (e.g., agglutinative or pro-drop languages), or non-Latin scripts requires separate evaluation. Multilingual encoder variants (e.g., mBERT, XLM-R) would be a natural starting point.

\paragraph{Mean ablation is off-manifold.}
Replacing selected neurons with their training-mean pushes activations off the natural data manifold, which can introduce artefacts unrelated to the detection signal~\citep{heimersheim2024patching}. The reverse-patching experiment (§\ref{sec:results-patching}) provides a complementary on-manifold necessity test using real donor activations; the mean-ablation result should be read alongside it rather than in isolation. We treat the combined evidence as a conservative lower bound on necessity.

\paragraph{Layer-asymmetry claim across two base generators.}
The observation that instruction-tuned generators concentrate 30--36\% of stable neurons in layer~12 while pure-base generators have $\leq$14\% is supported by two independent base models (GPT-2 and MPT-30B). However, the two base generators differ in their peak layer (GPT-2: L11; MPT-30B: L1/L8), so the claim is best stated as L12-avoidance rather than a strict L11-peaking rule. We only claim lower L12 mass in base models; the specific lower-layer peak is not consistent across base models.

\paragraph{Adversarial robustness not evaluated.}
RAID includes eleven adversarial rephrasing attacks (paraphrase, homoglyph, synonym substitution, etc.) that substantially reduce detector accuracy~\citep{dugan2024raid}. Whether the stable neurons identified here remain causal under these attacks is left to future work; the present analysis covers only the non-adversarial RAID subset.

\section*{Ethical considerations}
\label{sec:ethics}

AI-text detection is a dual-use capability: detectors can support platform integrity and journalistic verification, but can also be used to penalise legitimate AI-assisted writing or to guide adversarial evasion. This work is mechanistic-interpretability research: we study which neurons drive an existing frozen encoder's representations, not deploy a new detection system. All data come from the publicly released RAID benchmark \citep{dugan2024raid}; no human subjects were involved and no personally identifiable information was collected or processed. The stable neuron sets we identify could in principle inform evasion strategies (by targeting the identified neuron sets), but the same information could equally guide more robust detector design. We release all experiment code to support reproducibility (provided as supplementary material).

\bibliography{refs}

\appendix

\section{Pipeline Details}
\label{app:pipeline}

Figure~\ref{fig:pipeline} in Section~\ref{sec:introduction} provides a visual overview of the pipeline described below.

\paragraph{Data sampling.}
For each generator we draw $N = 7{,}500$ balanced samples from the RAID benchmark: equal numbers of AI-generated and human-written texts, stratified across the six RAID domains (abstracts, books, news, Reddit, reviews, Wikipedia) using stratified sampling without replacement. Texts shorter than 50 whitespace-delimited tokens are excluded. Human texts for each generator are the original source passages from which the AI texts were generated; this pairs each AI sample with a human sample from the same domain and source.

\paragraph{Activation extraction.}
We encode each text with \bertbase\ using HuggingFace Transformers~\citep{wolf2020transformers}. The input is truncated to 512 WordPiece tokens. We extract the CLS-token hidden state from \emph{all} 12 transformer layers and concatenate them into a single vector $\mathbf{z} \in \mathbb{R}^{9{,}216}$ ($12 \times 768$). Gradients are disabled throughout; BERT weights are never updated.

\paragraph{Normalisation.}
Before any probe training, each of the 9{,}216 neurons is $z$-scored with a \texttt{StandardScaler} fit on the training fold and applied to the test fold. The same scaler is reused for all downstream experiments (patching, ablation) within a given cell so that scale is held constant across interventions.

\paragraph{Cross-validation protocol.}
We use 5-fold stratified cross-validation repeated with 3 random seeds (42, 123, 456), giving 15 independent evaluation cells per generator. The stratification is over the binary AI/human label; domain balance is not enforced at the fold level but is approximately preserved because the initial sample is domain-stratified.

\paragraph{L1 selection.}
The \emph{selector} is a $\ell_1$-regularised logistic regression (\texttt{liblinear} solver, $C=0.005$, \texttt{max\_iter}$=1{,}000$, \texttt{class\_weight=balanced}) fit on the training fold. Concretely, \texttt{liblinear} minimises
\[
  \tfrac{1}{2C}\|\mathbf{w}\|_1
  + \sum_{i=1}^{n} \log\!\bigl(1 + \exp(-y_i\, \mathbf{x}_i^{\top}\mathbf{w})\bigr)
\]
over training-fold samples $(\mathbf{x}_i, y_i)$ with $y_i \in \{-1, +1\}$; smaller $C$ inflates the $\ell_1$ coefficient and therefore enforces a sparser solution. The selected set $\mathcal{S}$ for a given cell is the set of neurons with non-zero weight. All 9{,}216 neurons are provided as input; the regulariser drives most to zero.

\paragraph{Stable neuron aggregation.}
A neuron is \emph{stable} if it appears in $\mathcal{S}$ in at least 80\% of the 15 cells ($\geq 12$ out of 15). This threshold was chosen to require cross-seed and cross-fold consistency simultaneously: a neuron must survive in the large majority of both fold and seed resamplings, not merely in a bare majority of cells. Neurons appearing in 0--11 cells are discarded; those in 12--15 cells form the stable set $\mathcal{S}^{*}$.

\paragraph{L2 evaluation probe.}
The \emph{evaluator} is a separate $\ell_2$-regularised logistic regression (\texttt{lbfgs} solver, $C=1.0$, \texttt{max\_iter}$=1{,}000$) trained on each fold's training split using the same scaler. For all main experiments (sparse-probe accuracy, activation patching, mean ablation), it is trained once on all 9{,}216 neurons per fold and held frozen for the corresponding downstream interventions; the interventions modify input activations, not the probe. The restricted-probe analysis (Appendix~\ref{app:restricted}) is the only setting in which the L2 probe is re-trained on a reduced input dimension (the selected columns only).

\paragraph{Patching protocol.}
Patching uses same-domain donor matching: for each human target sample, a donor AI sample from the same domain is drawn without replacement. The selected $k$ neuron activations of the target are replaced with the corresponding values from the donor. The $k$ neurons are the first $k$ members of the per-fold candidate set $\mathcal{S}$, ranked by descending absolute L1 coefficient (most important first). For the full-$k$ sweep, all $|\mathcal{S}|$ per-fold neurons are patched (53--92 depending on generator and fold).

\paragraph{Software and licensing.}
We use HuggingFace Transformers v5.3.0~\citep{wolf2020transformers} for BERT inference and \texttt{scikit-learn} v1.8.0~\citep{scikit-learn} for all linear probes. RAID is released under the MIT license; \bertbase is released under the Apache 2.0 license. We use both within their permitted scope (non-commercial research analysis); no data or model weights are redistributed.

\paragraph{Computational budget.}
\bertbase has 110M parameters and is used in inference-only mode (no fine-tuning) throughout. CLS activations for all 45{,}000 samples are encoded once using the GPU (${\sim}30$ minutes) and cached; all downstream experiments (L1/L2 probes, stability sweep, mean ablation, bidirectional patching $k$-sweep, LOGO evaluation) operate on the cached representations. The complete pipeline runs in approximately 7--8 wall-clock hours on a laptop (AMD Ryzen 5 5600H, NVIDIA GeForce RTX 3060 Laptop GPU, 6\,GB VRAM). The remaining stages are all CPU-bound: the multi-generator stability grid takes ${\sim}1$ hour, the six per-generator main runs including the $k$-sweep with $n_{\text{shuffles}}=20$ donor permutations take ${\sim}5$ hours total, and the LOGO evaluation across five family folds takes ${\sim}15$ minutes. L1/L2 logistic regression, mean ablation, and activation patching with cached representations do not benefit from GPU acceleration. The pipeline is therefore I/O- rather than compute-bound after the initial encoding pass.

\section{Stability Sweep}
\label{app:stability}

Section~\ref{sec:setup-stability} justifies the choice of $C=0.005$ and $N=7{,}500$ by reference to this appendix. The stability sweep evaluates mean pairwise Jaccard similarity between selection sets produced by 15 independent draws of the L1 selector (varying random seed and fold assignment) for all six RAID generators across a grid of regularisation strengths $C \in \{0.001, 0.002, 0.005, 0.01, 0.02, 0.05\}$ and sample sizes $N \in \{500, 1000, 2000, 3500, 5000, 7500\}$ ($K=15$ draws per cell). Table~\ref{tab:stability-grid} reports the full grid. Higher Jaccard means the selection is more stable. Values in parentheses are the mean number of neurons selected per draw $\bar{n}$.

\begin{table*}[!htbp]
\centering
\small
\caption{Mean Jaccard stability (and mean $\bar{n}$ selected) across 15 random draws of the L1 selector. Rows are $C$ values; columns are $N$ values. Larger Jaccard is better. \textbf{Bold}: chosen operating point ($C=0.005$, $N=7{,}500$). Dagger ($\dagger$): degenerate cell ($\bar{n}<1$, Jaccard trivially 1.0 or unstable). $N=10{,}000$ omitted (only $K=1$ draw possible; pool exhausted).}
\label{tab:stability-grid}
\begin{tabular}{lrrrrrr}
\toprule
$C$ & $N=500$ & $N=1000$ & $N=2000$ & $N=3500$ & $N=5000$ & $N=7500$ \\
\midrule
\multicolumn{7}{l}{\textit{GPT-4}} \\
0.001 & $1.000^\dagger$ (0)  & $1.000^\dagger$ (0)  & $1.000^\dagger$ (0)  & 0.867 (0)   & 0.689 (8)   & 0.808 (18) \\
0.002 & $1.000^\dagger$ (0)  & $1.000^\dagger$ (0)  & 0.680 (3)   & 0.664 (16)  & 0.692 (28)  & 0.799 (44) \\
0.005 & $1.000^\dagger$ (0)  & 0.433 (9)   & 0.510 (29)  & 0.617 (51)  & 0.684 (65)  & \textbf{0.751 (80)} \\
0.01  & 0.314 (9)   & 0.356 (29)  & 0.488 (54)  & 0.553 (78)  & 0.598 (97)  & 0.681 (123) \\
0.02  & 0.254 (29)  & 0.355 (53)  & 0.425 (82)  & 0.460 (117) & 0.509 (143) & 0.625 (179) \\
0.05  & 0.205 (57)  & 0.282 (88)  & 0.310 (134) & 0.353 (184) & 0.414 (232) & 0.535 (283) \\
\midrule
\multicolumn{7}{l}{\textit{GPT-2}} \\
0.001 & $1.000^\dagger$ (0)  & $1.000^\dagger$ (0)  & $1.000^\dagger$ (0)  & $1.000^\dagger$ (0)  & 0.579 (2)   & 0.932 (14) \\
0.002 & $1.000^\dagger$ (0)  & $1.000^\dagger$ (0)  & $1.000^\dagger$ (0)  & 0.572 (12)  & 0.744 (19)  & 0.833 (37) \\
0.005 & $1.000^\dagger$ (0)  & 0.288 (3)   & 0.464 (20)  & 0.530 (49)  & 0.581 (70)  & \textbf{0.676 (103)} \\
0.01  & 0.197 (3)   & 0.308 (22)  & 0.385 (55)  & 0.443 (97)  & 0.512 (131) & 0.660 (183) \\
0.02  & 0.202 (21)  & 0.260 (53)  & 0.304 (106) & 0.405 (169) & 0.463 (216) & 0.574 (289) \\
0.05  & 0.133 (64)  & 0.191 (118) & 0.241 (204) & 0.306 (307) & 0.357 (388) & 0.489 (505) \\
\midrule
\multicolumn{7}{l}{\textit{MPT}} \\
0.001 & $1.000^\dagger$ (0)  & $1.000^\dagger$ (0)  & $1.000^\dagger$ (0)  & 0.467 (1)   & 0.861 (7)   & 0.840 (14) \\
0.002 & $1.000^\dagger$ (0)  & $1.000^\dagger$ (0)  & 0.546 (3)   & 0.704 (13)  & 0.774 (21)  & 0.832 (33) \\
0.005 & $1.000^\dagger$ (0)  & 0.496 (8)   & 0.619 (21)  & 0.624 (38)  & 0.638 (53)  & \textbf{0.707 (78)} \\
0.01  & 0.341 (7)   & 0.437 (22)  & 0.463 (42)  & 0.455 (74)  & 0.529 (99)  & 0.666 (130) \\
0.02  & 0.281 (21)  & 0.300 (43)  & 0.339 (84)  & 0.409 (125) & 0.448 (157) & 0.565 (210) \\
0.05  & 0.164 (52)  & 0.206 (90)  & 0.241 (151) & 0.294 (227) & 0.347 (296) & 0.477 (377) \\
\midrule
\multicolumn{7}{l}{\textit{Mistral}} \\
0.001 & $1.000^\dagger$ (0)  & $1.000^\dagger$ (0)  & $1.000^\dagger$ (0)  & $1.000^\dagger$ (0)  & 0.705 (7)   & 0.867 (19) \\
0.002 & $1.000^\dagger$ (0)  & $1.000^\dagger$ (0)  & 0.214 (1)   & 0.714 (17)  & 0.731 (27)  & 0.860 (33) \\
0.005 & $1.000^\dagger$ (0)  & 0.435 (8)   & 0.552 (26)  & 0.620 (38)  & 0.684 (52)  & \textbf{0.718 (78)} \\
0.01  & 0.259 (7)   & 0.461 (25)  & 0.450 (45)  & 0.475 (71)  & 0.524 (98)  & 0.632 (145) \\
0.02  & 0.322 (24)  & 0.317 (44)  & 0.333 (82)  & 0.357 (133) & 0.424 (178) & 0.546 (239) \\
0.05  & 0.171 (53)  & 0.190 (96)  & 0.220 (165) & 0.288 (248) & 0.345 (316) & 0.482 (415) \\
\midrule
\multicolumn{7}{l}{\textit{LLaMA}} \\
0.001 & $1.000^\dagger$ (0)  & $1.000^\dagger$ (0)  & $1.000^\dagger$ (0)  & $1.000^\dagger$ (0)  & 0.703 (7)   & 0.816 (22) \\
0.002 & $1.000^\dagger$ (0)  & $1.000^\dagger$ (0)  & 0.783 (2)   & 0.637 (18)  & 0.782 (30)  & 0.840 (39) \\
0.005 & $1.000^\dagger$ (0)  & 0.464 (7)   & 0.559 (30)  & 0.681 (44)  & 0.670 (53)  & \textbf{0.748 (78)} \\
0.01  & 0.291 (7)   & 0.444 (27)  & 0.519 (48)  & 0.541 (72)  & 0.618 (96)  & 0.696 (120) \\
0.02  & 0.280 (23)  & 0.380 (46)  & 0.406 (78)  & 0.452 (114) & 0.530 (147) & 0.616 (188) \\
0.05  & 0.208 (49)  & 0.259 (86)  & 0.291 (137) & 0.345 (201) & 0.417 (249) & 0.534 (310) \\
\midrule
\multicolumn{7}{l}{\textit{Cohere}} \\
0.001 & $1.000^\dagger$ (0)  & $1.000^\dagger$ (0)  & $1.000^\dagger$ (0)  & $1.000^\dagger$ (0)  & 0.692 (1)   & 0.814 (13) \\
0.002 & $1.000^\dagger$ (0)  & $1.000^\dagger$ (0)  & $1.000^\dagger$ (0)  & 0.568 (10)  & 0.708 (21)  & 0.762 (35) \\
0.005 & $1.000^\dagger$ (0)  & 0.137 (2)   & 0.469 (21)  & 0.553 (42)  & 0.612 (57)  & \textbf{0.715 (88)} \\
0.01  & 0.129 (2)   & 0.309 (19)  & 0.370 (50)  & 0.414 (85)  & 0.476 (117) & 0.579 (172) \\
0.02  & 0.191 (21)  & 0.211 (51)  & 0.263 (102) & 0.310 (164) & 0.364 (225) & 0.497 (327) \\
0.05  & 0.101 (61)  & 0.127 (122) & 0.165 (226) & 0.217 (352) & 0.285 (472) & 0.443 (642) \\
\bottomrule
\end{tabular}
\end{table*}

Four observations drive the choice of $C=0.005$, $N=7{,}500$. \textbf{(1) Sparse-end cut-off:} $C \leq 0.002$ yields high or trivial Jaccard at small $N$ but selects fewer than 22 neurons at $N=7{,}500$ across all generators: too sparse to support the interventions we run. Among operating points selecting $\geq 50$ neurons per draw, $C=0.005$ achieves the highest minimum-over-6-generators Jaccard. \textbf{(2) Regularisation:} at any fixed $N \geq 2{,}000$ in the $C \geq 0.005$ region, $C=0.005$ dominates all weaker regularisation values (larger $C$) on every generator. \textbf{(3) Sample size:} for $C=0.005$ Jaccard increases monotonically in $N$, with the largest gains between $N=1{,}000$ and $N=3{,}500$. The $N=5{,}000 \to 7{,}500$ step adds 0.016--0.103 per generator; $N=10{,}000$ cannot be evaluated because the pool is exhausted ($K=1$). The chosen operating point achieves a minimum-over-6-generators Jaccard of 0.676, with GPT-2 as the bottleneck (0.676) and the remaining five generators ranging 0.707--0.751. \textbf{(4) Degenerate region:} at $N \leq 500$ and $C=0.005$, every generator selects zero neurons (Jaccard $=1.0$ trivially); $N=7{,}500$ reliably produces non-empty sets across all six generators.

\section{Per-Cell Metrics}
\label{app:per-cell}

Table~\ref{tab:per-cell} reports L2-probe accuracy, AUC-ROC, F1, and number of L1-selected neurons for every evaluation cell (seed $\times$ fold).

\begin{table*}[!htbp]
\centering
\footnotesize
\caption{Per-cell L2-probe metrics for all six generators. Each block of 15 rows is one generator; columns are seed and fold index. Acc: accuracy; AUC: AUC-ROC; $n_{\text{sel}}$: neurons selected by L1 in this cell.}
\label{tab:per-cell}
\begin{minipage}[t]{0.48\textwidth}
\centering
\begin{tabular}{llrrrr}
\toprule
Gen. & Seed & Fold & Acc & AUC & $n_{\text{sel}}$ \\
\midrule
\multirow{15}{*}{GPT-4}
 & 42  & 0 & 0.9887 & 0.9991 & 74 \\
 & 42  & 1 & 0.9853 & 0.9992 & 73 \\
 & 42  & 2 & 0.9887 & 0.9991 & 80 \\
 & 42  & 3 & 0.9853 & 0.9989 & 74 \\
 & 42  & 4 & 0.9927 & 0.9989 & 70 \\
 & 123 & 0 & 0.9860 & 0.9987 & 70 \\
 & 123 & 1 & 0.9880 & 0.9992 & 76 \\
 & 123 & 2 & 0.9860 & 0.9989 & 73 \\
 & 123 & 3 & 0.9907 & 0.9995 & 73 \\
 & 123 & 4 & 0.9867 & 0.9991 & 75 \\
 & 456 & 0 & 0.9907 & 0.9989 & 77 \\
 & 456 & 1 & 0.9907 & 0.9994 & 75 \\
 & 456 & 2 & 0.9907 & 0.9991 & 72 \\
 & 456 & 3 & 0.9840 & 0.9981 & 71 \\
 & 456 & 4 & 0.9847 & 0.9993 & 79 \\
\midrule
\multirow{15}{*}{GPT-2}
 & 42  & 0 & 0.9587 & 0.9923 & 80 \\
 & 42  & 1 & 0.9587 & 0.9907 & 81 \\
 & 42  & 2 & 0.9593 & 0.9906 & 80 \\
 & 42  & 3 & 0.9593 & 0.9918 & 82 \\
 & 42  & 4 & 0.9613 & 0.9909 & 90 \\
 & 123 & 0 & 0.9627 & 0.9912 & 90 \\
 & 123 & 1 & 0.9593 & 0.9912 & 82 \\
 & 123 & 2 & 0.9647 & 0.9931 & 84 \\
 & 123 & 3 & 0.9513 & 0.9908 & 86 \\
 & 123 & 4 & 0.9613 & 0.9916 & 80 \\
 & 456 & 0 & 0.9573 & 0.9929 & 90 \\
 & 456 & 1 & 0.9580 & 0.9914 & 84 \\
 & 456 & 2 & 0.9547 & 0.9911 & 82 \\
 & 456 & 3 & 0.9620 & 0.9931 & 92 \\
 & 456 & 4 & 0.9580 & 0.9902 & 87 \\
\midrule
\multirow{15}{*}{MPT}
 & 42  & 0 & 0.9700 & 0.9939 & 59 \\
 & 42  & 1 & 0.9720 & 0.9931 & 60 \\
 & 42  & 2 & 0.9733 & 0.9954 & 60 \\
 & 42  & 3 & 0.9700 & 0.9947 & 61 \\
 & 42  & 4 & 0.9800 & 0.9960 & 61 \\
 & 123 & 0 & 0.9687 & 0.9908 & 56 \\
 & 123 & 1 & 0.9700 & 0.9961 & 68 \\
 & 123 & 2 & 0.9780 & 0.9951 & 60 \\
 & 123 & 3 & 0.9700 & 0.9934 & 65 \\
 & 123 & 4 & 0.9700 & 0.9949 & 60 \\
 & 456 & 0 & 0.9740 & 0.9938 & 65 \\
 & 456 & 1 & 0.9700 & 0.9941 & 66 \\
 & 456 & 2 & 0.9740 & 0.9939 & 59 \\
 & 456 & 3 & 0.9747 & 0.9942 & 61 \\
 & 456 & 4 & 0.9680 & 0.9931 & 59 \\
\bottomrule
\end{tabular}
\end{minipage}%
\hfill
\begin{minipage}[t]{0.48\textwidth}
\centering
\begin{tabular}{llrrrr}
\toprule
Gen. & Seed & Fold & Acc & AUC & $n_{\text{sel}}$ \\
\midrule
\multirow{15}{*}{Mistral}
 & 42  & 0 & 0.9727 & 0.9952 & 59 \\
 & 42  & 1 & 0.9700 & 0.9950 & 53 \\
 & 42  & 2 & 0.9607 & 0.9946 & 62 \\
 & 42  & 3 & 0.9713 & 0.9954 & 56 \\
 & 42  & 4 & 0.9780 & 0.9961 & 58 \\
 & 123 & 0 & 0.9720 & 0.9945 & 60 \\
 & 123 & 1 & 0.9667 & 0.9929 & 61 \\
 & 123 & 2 & 0.9753 & 0.9966 & 55 \\
 & 123 & 3 & 0.9700 & 0.9939 & 55 \\
 & 123 & 4 & 0.9747 & 0.9954 & 62 \\
 & 456 & 0 & 0.9733 & 0.9967 & 60 \\
 & 456 & 1 & 0.9667 & 0.9935 & 54 \\
 & 456 & 2 & 0.9747 & 0.9949 & 60 \\
 & 456 & 3 & 0.9693 & 0.9961 & 57 \\
 & 456 & 4 & 0.9660 & 0.9937 & 55 \\
\midrule
\multirow{15}{*}{LLaMA}
 & 42  & 0 & 0.9920 & 0.9982 & 63 \\
 & 42  & 1 & 0.9833 & 0.9973 & 64 \\
 & 42  & 2 & 0.9900 & 0.9989 & 63 \\
 & 42  & 3 & 0.9813 & 0.9990 & 64 \\
 & 42  & 4 & 0.9827 & 0.9981 & 64 \\
 & 123 & 0 & 0.9900 & 0.9995 & 65 \\
 & 123 & 1 & 0.9847 & 0.9970 & 63 \\
 & 123 & 2 & 0.9840 & 0.9988 & 62 \\
 & 123 & 3 & 0.9880 & 0.9993 & 61 \\
 & 123 & 4 & 0.9853 & 0.9980 & 64 \\
 & 456 & 0 & 0.9840 & 0.9977 & 67 \\
 & 456 & 1 & 0.9853 & 0.9985 & 67 \\
 & 456 & 2 & 0.9853 & 0.9986 & 62 \\
 & 456 & 3 & 0.9900 & 0.9991 & 61 \\
 & 456 & 4 & 0.9867 & 0.9985 & 64 \\
\midrule
\multirow{15}{*}{Cohere}
 & 42  & 0 & 0.9107 & 0.9716 & 71 \\
 & 42  & 1 & 0.8967 & 0.9673 & 72 \\
 & 42  & 2 & 0.8960 & 0.9612 & 72 \\
 & 42  & 3 & 0.9213 & 0.9710 & 73 \\
 & 42  & 4 & 0.9053 & 0.9669 & 72 \\
 & 123 & 0 & 0.8973 & 0.9632 & 65 \\
 & 123 & 1 & 0.9060 & 0.9665 & 76 \\
 & 123 & 2 & 0.9133 & 0.9692 & 72 \\
 & 123 & 3 & 0.9013 & 0.9704 & 76 \\
 & 123 & 4 & 0.9027 & 0.9681 & 72 \\
 & 456 & 0 & 0.9107 & 0.9704 & 69 \\
 & 456 & 1 & 0.9013 & 0.9671 & 73 \\
 & 456 & 2 & 0.8980 & 0.9629 & 79 \\
 & 456 & 3 & 0.9173 & 0.9729 & 70 \\
 & 456 & 4 & 0.9000 & 0.9653 & 73 \\
\bottomrule
\end{tabular}
\end{minipage}
\end{table*}

\section{AUC-ROC vs.\ L1 Selection}
\label{app:auc-l1}

In this appendix, AUC-ROC serves as a univariate neuron-ranking criterion: each neuron is scored by its individual AUC-ROC against the binary label on the training fold, rather than as a probe performance metric.

We compare L1-regularised selection (the method used throughout this paper) to a univariate AUC-ROC-based ranking: for each neuron, we compute its individual AUC-ROC against the binary label on the training fold, together with a Mann--Whitney $U$ test with Bonferroni correction at $\alpha=0.001$; retain neurons that are both statistically significant and exceed a per-neuron AUC-ROC threshold of $0.7$ (or fall below $0.3$). Selected set sizes therefore vary by generator rather than being constrained to match the L1 count. We evaluate the informativeness of each selected set by \emph{complement ablation}: the L2 probe is re-evaluated after mean-ablating all neurons \emph{not} in the selected set (i.e., only the selected neurons are retained). A smaller drop from the full-probe baseline indicates the selected set is more self-contained.

\begin{table}[!htbp]
\centering
\small
\caption{L1 vs.\ AUC-ROC selection: mean across 15 cells. $n$: neurons selected. Jaccard: overlap between the two selection sets. Drop: accuracy drop from baseline when complement is ablated (positive = degradation). Three regimes are visible: parsimony (GPT-4, MPT, Mistral, LLaMA: L1 selects far fewer neurons with equivalent or smaller drop), convergence (GPT-2, similar $n$ and similar drop), and reversal (Cohere: AUC-ROC selects fewer with smaller drop).}
\label{tab:auc-l1}
\resizebox{\columnwidth}{!}{%
\begin{tabular}{lrrrrr}
\toprule
Gen. & $n_{\text{AUC}}$ & $n_{\text{L1}}$ & Jaccard & AUC drop & L1 drop \\
\midrule
GPT-4   & 491 & 74  & 0.127 & +1.02pp & +0.03pp \\
GPT-2   & 86  & 85  & 0.096 & +0.06pp & +0.11pp \\
MPT     & 784 & 61  & 0.054 & +6.90pp & +0.05pp \\
Mistral & 238 & 58  & 0.148 & +0.52pp & +0.10pp \\
LLaMA   & 308 & 64  & 0.139 & +0.55pp & +0.11pp \\
Cohere  & 38  & 72  & 0.185 & +0.28pp & +1.06pp \\
\bottomrule
\end{tabular}}
\end{table}

The Jaccard overlap between L1 and AUC-ROC-selected sets is low (0.05--0.19; Table~\ref{tab:auc-l1}) despite both being applied to the same data. The two criteria thus select largely non-overlapping neurons. For GPT-4, MPT, Mistral, and LLaMA, L1 achieves equivalent or better task performance with far fewer neurons ($4\times$ to $13\times$ fewer): ablating L1's 58--74 neurons costs only 0.03--0.11pp, while ablating AUC-ROC's 238--784 neurons costs 0.52--6.90pp. This parsimony makes L1 preferable for mechanistic analysis: a smaller identified set is easier to interpret and less likely to include redundant or near-duplicated neurons.

For GPT-2, the two methods select nearly the same number of neurons (85 vs.\ 86) with similar ablation cost; neither dominates. For Cohere, AUC-ROC selects \emph{fewer} neurons (38 vs.\ 72) and its ablation cost is lower (+0.28pp vs.\ +1.06pp): Cohere's detection signal is concentrated in a small number of individually highly-discriminative neurons that univariate ranking recovers more directly than joint L1 regularisation. This reversal fits Cohere's distinct geometry (§\ref{sec:results-necessity}, Appendix~\ref{app:cav}): the model with the lowest baseline probe accuracy is also the one where AUC-ROC beats L1 on parsimony.

\section{Signal Localisation: Restricted Probe}
\label{app:restricted}

The necessity result (§\ref{sec:results-necessity}) uses the full-feature L2 probe, which retains thousands of redundant correlates. A complementary question is whether the stable neurons alone are sufficient for a \textit{freshly trained} probe, and how this compares to ablating the complement. We measure the \textit{localisation gap}: the accuracy difference between a probe retrained on only the $|\mathcal{S}^{*}|$ stable neurons (\textit{smaller model}) and the original L2 probe evaluated with non-stable neurons mean-ablated (\textit{ablate complement}). A large positive gap means the stable set can support an independent probe on its own, but the full-feature probe leaned on non-stable neurons to define its decision boundary.

\begin{table}[!htbp]
\centering
\resizebox{\columnwidth}{!}{%
\begin{tabular}{lrr}
\toprule
Generator & Smaller model & Gap vs ablate-complement \\
\midrule
GPT-4   & $0.972\pm0.005$ & $+0.41 \pm 0.42$\,pp \\
GPT-2   & $0.913\pm0.006$ & $+3.63 \pm 0.81$\,pp \\
MPT     & $0.944\pm0.005$ & $+4.68 \pm 1.14$\,pp \\
Mistral & $0.934\pm0.005$ & $+0.93 \pm 0.42$\,pp \\
LLaMA   & $0.960\pm0.003$ & $+1.14 \pm 0.43$\,pp \\
Cohere  & $0.865\pm0.009$ & $+4.43 \pm 1.02$\,pp \\
\bottomrule
\end{tabular}%
}
\caption{Restricted probe results. \textit{Smaller model}: L2 probe retrained on only the $|\mathcal{S}^{*}|$ stable neurons. \textit{Gap}: smaller-model accuracy minus ablate-complement accuracy (positive = stable neurons self-sufficient; larger = more dependence on non-stable neurons in the full-feature probe). 15-cell means $\pm$ std.}
\label{tab:restricted-probe}
\end{table}

The pattern (Table~\ref{tab:restricted-probe}) splits cleanly along the base/instruction-tuned axis. Instruction-tuned generators (GPT-4, Mistral, LLaMA) show gaps of $\leq$1.2\,pp: their stable neurons are nearly self-sufficient and the full-feature probe does not rely heavily on non-stable dimensions. The two pure-base generators and Cohere show gaps of 3.6--4.7\,pp: their full-feature probes lean more on non-stable neurons, making complement ablation more costly. This points to a more distributed encoding structure for these three generators, in line with the layer-12 avoidance seen in base generators and the large ablation sensitivity of Cohere documented in the main body.

\section{Detection Signal Geometry (CAV Diagnostics)}
\label{app:cav}

To characterise the geometry of the detection signal within each stable set we compute a concept activation vector (CAV)~\citep{kim2018tcav} as the mean difference between AI and human activations restricted to the stable neurons, and compare it to the L1 probe weight vector. Table~\ref{tab:cav} reports cosine similarity between the CAV and the probe weight, logistic-regression accuracy on the CAV projection, and the norms of both vectors.

Cohere has the lowest CAV cosine similarity ($0.057 \pm 0.001$), the lowest LR accuracy on the CAV projection ($0.910 \pm 0.008$), and by far the largest probe weight norm ($108.3 \pm 2.1$); the detection signal is diffuse and the probe compensates for it with large weights. MPT-30B is a notable exception: it has a low cosine ($0.065 \pm 0.000$, close to Cohere's) but the \textit{largest} mean-difference norm ($\|\Delta\mu\| = 9.52$). The AI\,$-$\,human class separation is therefore geometrically strong per-axis yet misaligned with the probe direction. GPT-2 also shows a low cosine ($0.073 \pm 0.002$) and elevated weight norm ($69.9 \pm 2.9$), placing both base generators at the low-cosine end.

Instruction-tuned generators cluster at higher cosine ($0.110$--$0.138$) with the smallest weight norms. GPT-4 has the most compact weight vector ($45.8 \pm 1.7$); its detection signal is both geometrically aligned and concentrated. Each metric is mean\,$\pm$\,std across three LR train/test split seeds; the mean-difference norm is deterministic (no split randomness). These per-generator geometric differences track the intervention asymmetries documented in §\ref{sec:results-causal}.

\begin{table}[!htbp]
\centering
\resizebox{\columnwidth}{!}{%
\begin{tabular}{lrrrr}
\toprule
Generator & Cosine & LR acc & $\|\Delta\mu\|$ & $\|w\|$ \\
\midrule
GPT-4   & $0.138 \pm 0.004$ & $0.988 \pm 0.003$ &  7.34 & $\mathbf{45.8} \pm 1.7$ \\
GPT-2   & $0.073 \pm 0.002$ & $0.963 \pm 0.006$ &  6.84 & $69.9 \pm 2.9$ \\
MPT     & $0.065 \pm 0.000$ & $0.973 \pm 0.001$ & $\mathbf{9.52}$ & $53.8 \pm 2.7$ \\
Mistral & $0.110 \pm 0.001$ & $0.965 \pm 0.003$ &  6.38 & $59.2 \pm 2.3$ \\
LLaMA   & $0.128 \pm 0.001$ & $0.987 \pm 0.003$ &  6.57 & $48.3 \pm 0.9$ \\
Cohere  & $\mathbf{0.057} \pm 0.001$ & $\mathbf{0.910} \pm 0.008$ &  4.69 & $108.3 \pm 2.1$ \\
\bottomrule
\end{tabular}%
}
\caption{CAV diagnostics on per-generator stable sets. All values mean\,$\pm$\,std across 3 random seeds. Cosine: similarity between mean-difference CAV and L1 probe weight. LR acc: logistic-regression accuracy on the CAV projection. $\|\Delta\mu\|$: AI\,$-$\,human mean-difference norm (deterministic; bold\,=\,largest). $\|w\|$: L1 weight norm (bold\,=\,smallest).}
\label{tab:cav}
\end{table}

\section{Per-Domain Patching Breakdown}
\label{app:patching-perdomain}

Table~\ref{tab:perdomain} reports selected flip rates (\%) at $k = \text{full}$ broken down by RAID domain, averaged over all 15 cells (5 folds $\times$ 3 seeds). Values are 15-cell means; per-cell standard deviations ranged 0.6--2.6\,pp across all generator--domain combinations.

\begin{table*}[!htbp]
\centering
\small
\begin{tabular}{lrrrrrr}
\toprule
Domain & GPT-4 & GPT-2 & MPT & Mistral & LLaMA & Cohere \\
\midrule
Abstracts & 1.44 & 3.08 & 1.39 & 1.79 & 1.47 &  9.89 \\
Books     & 1.08 & 1.93 & 1.47 & 0.92 & 1.21 & 10.06 \\
News      & 1.61 & 4.92 & 2.74 & 2.11 & 1.01 &  9.67 \\
Reddit    & 1.05 & 3.22 & 1.36 & 1.34 & 0.76 &  7.50 \\
Reviews   & 0.97 & 1.96 & 0.55 & 0.62 & 1.14 &  5.48 \\
Wiki      & 1.25 & 3.65 & 1.61 & 3.54 & 0.83 &  6.26 \\
\midrule
Mean (full) & 1.23 & 3.13 & 1.52 & 1.73 & 1.07 & 8.15 \\
\midrule
CV          & 0.18 & 0.33 & 0.42 & 0.55 & 0.22 & 0.22 \\
\bottomrule
\end{tabular}
\caption{Per-domain selected flip rate (\%) at $k = \text{full}$, 15-cell means. Domain means may differ slightly from the overall mean due to variation in the number of valid-human pairs per domain. GPT-2 shows elevated flip rates on news, Reddit, and wiki (3.1--4.9\%) relative to books and reviews (1.9--2.0\%). Cohere is uniformly high across all six domains (5.5--10.1\%); its probe operates near the decision boundary throughout. Mistral shows a notable wiki spike (3.54\%) while reviews are weak (0.62\%). CV: coefficient of variation over each generator's six domain flip rates. See §\ref{sec:patching-results} for interpretation.}
\label{tab:perdomain}
\end{table*}

\paragraph{Domain sensitivity across generators.}
The bottom row of Table~\ref{tab:perdomain} summarises each generator's spread
across domains as a coefficient of variation. Mistral (0.55), MPT (0.42), and
GPT-2 (0.33) are domain-sensitive; GPT-4 (0.18), LLaMA (0.22), and Cohere
(0.22) are close to domain-invariant. Cohere's flip rates are high everywhere,
which is consistent with a probe operating near its decision boundary
regardless of content, whereas GPT-2 separates news, Reddit, Wikipedia, and
abstracts (3.1--4.9\%) from books and reviews (1.9--2.0\%). Both pure-base
generators fall in the sensitive group, but Mistral does as well, so domain
sensitivity does not follow the base/instruction-tuned split that organises the
layer distributions (§\ref{sec:crossgen-layers}). The two axes are related but
distinct, and the present data do not identify what drives the difference.

\section{Stable-Neuron Layer Distributions (Full Table)}
\label{app:layer-dist}

Table~\ref{tab:layer-dist} reports the per-layer distribution of stable neurons for all six generators, summarised in §\ref{sec:crossgen-layers}.

\begin{table}[!htbp]
\centering
\resizebox{\columnwidth}{!}{%
\begin{tabular}{lrrrrrr}
\toprule
Layer & GPT-4 & GPT-2 & MPT & Mistral & LLaMA & Cohere \\
\midrule
1--10  & 36 & 41 & \textbf{44} & 24 & 30 & 40 \\
11     &  6 & \textbf{11} &  6 &  5 &  2 &  3 \\
12     & \textbf{18} &  8 &  2 & \textbf{16} & \textbf{16} & \textbf{19} \\
\midrule
\% in L12 & 30.0 & 13.3 & 3.8 & 35.6 & 33.3 & 30.6 \\
\bottomrule
\end{tabular}%
}
\caption{Stable-neuron counts by layer group across six generators. Bold indicates the peak layer group for each generator. Instruction-tuned generators (GPT-4, Mistral, LLaMA, Cohere) concentrate 30--36\% of stable neurons in layer~12; both pure-base generators (GPT-2, MPT) have $\leq$14\% in layer~12.}
\label{tab:layer-dist}
\end{table}

\section{Pairwise Jaccard Similarity (Full Table)}
\label{app:jaccard}

\begin{figure}[!htbp]
\centering
\includegraphics[width=\columnwidth]{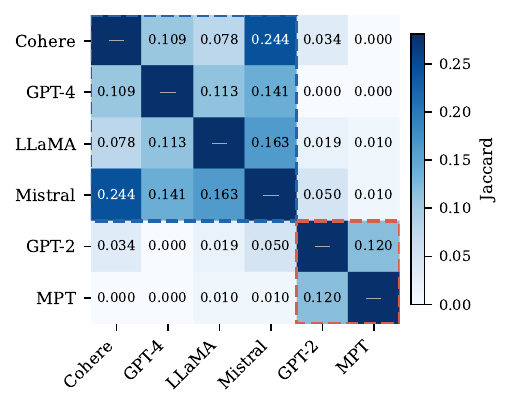}
\caption{Pairwise Jaccard similarity heatmap (6$\times$6). Generators reordered: instruction-tuned (IT) first, base last. Dashed blue box: IT$\times$IT block. Dashed red box: base$\times$base block. The bipartite structure (higher within-group similarity, near-zero cross-group overlap) is immediately visible.}
\label{fig:jaccard}
\end{figure}

Table~\ref{tab:jaccard} reports pairwise Jaccard similarity between stable sets for all 15 generator pairs, discussed in §\ref{sec:crossgen-jaccard}. Figure~\ref{fig:jaccard} visualises the same data as a heatmap with generators reordered to expose the base/instruction-tuned partition.

\begin{table}[!htbp]
\centering
\small
\begin{tabular}{lrr}
\toprule
Pair & Jaccard & $\times$ chance \\
\midrule
Cohere -- Mistral & 0.244 & 86$\times$ \\
LLaMA -- Mistral  & 0.163 & 64$\times$ \\
GPT-4 -- Mistral  & 0.141 & 51$\times$ \\
GPT-4 -- LLaMA    & 0.113 & 39$\times$ \\
\textbf{MPT -- GPT-2}      & \textbf{0.120} & \textbf{40$\times$} \\
Cohere -- GPT-4   & 0.109 & 33$\times$ \\
Cohere -- LLaMA   & 0.078 & 27$\times$ \\
GPT-2 -- Mistral  & 0.050 & 18$\times$ \\
Cohere -- GPT-2   & 0.034 & 10$\times$ \\
GPT-2 -- LLaMA    & 0.019 &  7$\times$ \\
MPT -- Mistral    & 0.010 &  4$\times$ \\
MPT -- LLaMA      & 0.010 &  4$\times$ \\
GPT-2 -- GPT-4    & 0.000 &  0$\times$ \\
MPT -- GPT-4      & 0.000 &  0$\times$ \\
MPT -- Cohere     & 0.000 &  0$\times$ \\
\midrule
Mean              & 0.073 & $\sim$24$\times$ \\
\bottomrule
\end{tabular}
\caption{Pairwise Jaccard similarity between stable sets (15 generator pairs). $\times$\,chance is observed Jaccard divided by the analytic random-null Jaccard for two sets of the same sizes drawn from 9{,}216 neurons ($\approx$0.003 per pair). The mpt--gpt2 pair (bold) is the highest base$\times$base overlap; all base$\times$instruction-tuned pairs cluster at the bottom.}
\label{tab:jaccard}
\end{table}

\section{Full $k$-Sweep Flip Rates}
\label{app:patching-sweep}

Table~\ref{tab:patching-sweep} reports selected flip rates (\%) across all $k$ values for each generator, referenced in §\ref{sec:patching-results}.

\begin{table*}[!htbp]
\centering
\small
\begin{tabular}{lrrrrrr}
\toprule
$k$ & GPT-4 & GPT-2 & MPT & Mistral & LLaMA & Cohere \\
\midrule
\multicolumn{7}{l}{\textit{Forward direction (AI donor $\to$ human target) --- selected $k$-sweep}} \\
1    & 0.02 & 0.05 & 0.04 & 0.06 & 0.07 & 0.13 \\
5    & 0.09 & 0.17 & 0.14 & 0.20 & 0.14 & 0.58 \\
10   & 0.15 & 0.29 & 0.25 & 0.34 & 0.27 & 0.99 \\
20   & 0.35 & 0.58 & 0.47 & 0.56 & 0.49 & 2.02 \\
50   & 0.80 & 1.71 & 1.13 & 1.48 & 0.86 & 5.43 \\
selected (full) & $1.23_{\pm 0.33}$ & $3.13_{\pm 0.54}$ & $1.52_{\pm 0.21}$ & $1.73_{\pm 0.27}$ & $1.07_{\pm 0.37}$ & $8.15_{\pm 1.27}$ \\
random (full)   & 0.08 & 0.32 & 0.14 & 0.12 & 0.11 & 0.79 \\
ratio           & 16$\times$ & 10$\times$ & 11$\times$ & 14$\times$ & 10$\times$ & 10$\times$ \\
\midrule
\multicolumn{7}{l}{\textit{Reverse direction (human donor $\to$ AI target) --- selected $k$-sweep}} \\
1    & 0.02 & 0.08 & 0.04 & 0.07 & 0.03 & 0.09 \\
5    & 0.08 & 0.16 & 0.13 & 0.23 & 0.08 & 0.55 \\
10   & 0.14 & 0.27 & 0.22 & 0.34 & 0.17 & 0.93 \\
20   & 0.21 & 0.57 & 0.43 & 0.49 & 0.36 & 1.61 \\
50   & 0.43 & 1.44 & 1.09 & 1.18 & 0.75 & 3.99 \\
selected (full) & $0.65_{\pm 0.21}$ & $2.41_{\pm 0.62}$ & $1.36_{\pm 0.40}$ & $1.35_{\pm 0.27}$ & $0.92_{\pm 0.21}$ & $5.74_{\pm 0.87}$ \\
random (full)   & 0.07 & 0.33 & 0.17 & 0.14 & 0.08 & 0.77 \\
ratio           & 9.3$\times$ & 7.3$\times$ & 8.2$\times$ & 9.7$\times$ & 11.9$\times$ & 7.4$\times$ \\
\bottomrule
\end{tabular}
\caption{Flip rate (\%) for both patching directions, 15-cell means (5 folds $\times$ 3 seeds). $k = \text{full}$ equals each cell's actual $|\mathcal{S}|$ (53--92 neurons); std shown only for selected $k = \text{full}$. \textbf{Forward direction} (AI donor $\to$ human target): selected flip rate is monotonically increasing in $k$ for every generator. Selected/random ratios at $k = \text{full}$ are 10--16$\times$ (precise values 9.7--15.7$\times$; see Table~\ref{tab:patching-full}). \textbf{Reverse direction} (human donor $\to$ AI target) is likewise monotonically increasing in $k$ for every generator; selected/random ratios at $k = \text{full}$ are 7--12$\times$, and absolute rates are below the forward direction at every $k$. See §\ref{sec:patching-results} for the partial-redundancy reading.}
\label{tab:patching-sweep}
\end{table*}

\section{What the Selected Neurons Track}
\label{app:features}

Section~\ref{sec:results-crossgen} characterises \textit{where} the stable
neurons sit; this appendix examines what they respond to. We correlate neuron
activations with four surface statistics of the input text: type--token ratio,
mean word length, mean sentence length, and punctuation density. These
statistics are simple and model-independent; the question they address is how
much of a stable neuron's behaviour a single surface cue accounts for.

\paragraph{Protocol.}
For each generator we rank its stable set $\mathcal{S}^{*}$ by mean $|w|$ in
the L1 selection probe, averaged over the 15 cells, and keep the ten
highest-weight neurons. Each neuron's activation is Spearman-correlated with
each of the four features over all $N=9{,}996$ cached samples, giving 40 tests
per generator, to which we apply a Benjamini--Hochberg correction at $q=0.05$
within the generator. Feature values are computed on the token sequence the
encoder actually saw, reconstructed by decoding the cached inputs, so features
and activations are index-aligned by construction.

\paragraph{Results.}
Table~\ref{tab:features} reports the strongest surviving association per
generator. The associations are consistent but partial: 31--37 of the 40 tests
survive correction for every generator, while the strongest effect per
generator is $|\rho| = 0.41$--$0.54$, or roughly 16--29\% of variance
explained. Most of the variance in these neurons is therefore unaccounted for
by all four features combined, and the selected set does not reduce to a single
surface cue.

Mean word length is the top-associated feature for five of six generators
(type--token ratio for MPT), but the \textit{sign} of that association is
inconsistent: positive for GPT-4, GPT-2, and LLaMA, negative for Mistral and
Cohere. A neuron whose activation increases with longer words under one
generator decreases with them under another. This is consistent with the
largely disjoint stable sets of §\ref{sec:crossgen-jaccard}, and it argues
against reading these neurons as a generator-independent detector of lexical
complexity.

The neurons carrying the strongest surface association are also not always the
layer-12 neurons that dominate the instruction-tuned distributions. They sit in
layer~12 for GPT-4 and LLaMA, but in layers~1 and~5 for GPT-2, MPT, Mistral,
and Cohere. Two neurons recur across generators (L12/308 for GPT-4 and LLaMA;
L5/531 for Mistral and Cohere), and both belong to the 17-neuron
cross-generator core of §\ref{sec:crossgen-core}.

Surface lexical statistics are thus measurably present in the selected subspace
and are the strongest single-feature account we obtained, but they remain an
incomplete one. Identifying the remaining structure would require learned
feature dictionaries rather than hand-chosen statistics, which we leave to
future work (see \hyperref[sec:limitations]{Limitations}).

\begin{table}[!htbp]
\centering
\small
\begin{tabular}{llrrr}
\toprule
Generator & Feature & $\rho$ & $\rho^2$ & Neuron \\
\midrule
GPT-4   & word length & $+0.405$ & 0.164 & L12/308 \\
GPT-2   & word length & $+0.537$ & 0.288 & L1/661  \\
MPT     & type--token & $-0.478$ & 0.229 & L1/188  \\
Mistral & word length & $-0.510$ & 0.261 & L5/531  \\
LLaMA   & word length & $+0.441$ & 0.195 & L12/308 \\
Cohere  & word length & $-0.454$ & 0.207 & L5/531  \\
\bottomrule
\end{tabular}
\caption{Strongest surviving neuron--feature association per generator, over
the ten highest-weight stable neurons $\times$ four surface features (40
Spearman tests per generator, Benjamini--Hochberg at $q=0.05$; 31--37 tests
survive per generator). $\rho^2$ is the approximate share of variance
explained. ``word length'' is mean word length, ``type--token'' the
type--token ratio. Neuron is given as layer/unit. All associations are moderate
and none approaches full localisation of the detection signal to a surface
cue.}
\label{tab:features}
\end{table}

\section{A Worked Patching Example}
\label{app:worked-example}

The flip rates of §\ref{sec:patching-results} are aggregates over 15 cells and
20 donor permutations. This appendix reports a single intervention in full.

We take the Cohere generator, seed 42, fold 0, and patch the full stable set
$\mathcal{S}^{*}$ (62 neurons, of which 19, or 31\%, lie in layer~12,
matching the 30--36\% band reported for instruction-tuned generators in
Figure~\ref{fig:layer-dist}). The target is a human-written news article; the
donor is a Cohere-generated news article from the same test fold, so the
same-domain constraint of §\ref{sec:patching-protocol} is respected. Both are
roughly 1{,}030 characters long.

Before patching, the frozen L2 probe classifies the target as human with
$P(\text{AI}) = 0.007$, a confident prediction. Replacing the 62 selected
activations with the donor's values, and leaving the other 9{,}154 neurons
unchanged, moves the probe to $P(\text{AI}) = 0.654$, so the prediction flips
to AI. Editing 0.67\% of the representation is sufficient to reverse a decision
the probe held with high confidence.

The same donor draw flips 47 of 750 valid targets (6.3\%), close to Cohere's
reported forward flip rate of 8.15\% at $k = \text{full}$
(Table~\ref{tab:patching-full}), so the example is representative rather than
extreme. Most targets do \textit{not} flip, which is the redundancy reported
in §\ref{sec:results-necessity}, observed at the level of a single draw.

\end{document}